\documentclass[sigconf,screen]{acmart}
\AtBeginDocument{%
  }

\setcopyright{acmlicensed}
\copyrightyear{2025}
\acmYear{2025}
\acmDOI{XXXXXXX.XXXXXXX}
\acmConference[MM '25]{Make sure to enter the correct
  conference title from your rights confirmation email}{October 27--31,
  2025}{Dublin, Ireland}
\acmISBN{978-x-xxxx-xxxx-x/YY/MM}

\acmSubmissionID{xxxx}

\begin{document}

\title{GSVR: 2D Gaussian-based Video Representation for 800+ FPS with Hybrid Deformation Field}
% GSRV: Representation for Videos via 2D Gaussian  and Disentangled Static-Dynamic Deformation Field
% GS-NeRV: Neural Video Representation with 2DGS and Deformation Field


% \affiliation{%

\author{Zhizhuo Pang}
\affiliation{\institution{College of Intelligence and Computing, Tianjin University} \country{China}} % 首次定义机构

\author{Zhihui Ke}
\affiliation{\institution{College of Intelligence and Computing, Tianjin University} \country{China}} % 用符号关联到同一机构

\author{Xiaobo Zhou}
\affiliation{\institution{College of Intelligence and Computing, Tianjin University} \country{China}} % 同上

\author{Tie Qiu}
\affiliation{\institution{College of Intelligence and Computing, Tianjin University} \country{China}} % 同上

% \author{Zhizhuo Pang}
% \affiliation{\institution{College of Intelligence and Computing, Tianjin University}} % 首次定义机构

% \author{Zhihui Ke}
% \affiliation{*} % 使用符号*引用第一个机构

% \author{Xiaobo Zhou}
% \affiliation{*} % 同上

% \author{Tie Qiu}
% \affiliation{*} % 同上
% \affiliation{%

% \affiliation{%

% \affiliation{%



%  1600 words better
% 摘要里应该讲一下三平面和多项式运动的
\begin{abstract}
Implicit neural representations for video have been recognized as a novel and promising form of video representation. Existing works pay more attention to improving video reconstruction quality but little attention to the decoding speed. However, the high computation of convolutional network used in existing methods leads to low decoding speed. Moreover, these convolution-based video representation methods also suffer from long training time, about 14 seconds  per frame to achieve 35+ PSNR on Bunny. To solve the above problems, we propose GSVR, a novel 2D Gaussian-based video representation, which achieves 800+ FPS and 35+ PSNR on Bunny, only needing a training time of $2$ seconds per frame. Specifically, we propose a hybrid deformation field to model the dynamics of the video, which combines two motion patterns, namely the tri-plane motion and the polynomial motion, to deal with the coupling of camera motion and object motion in the video. Furthermore, we propose a Dynamic-aware Time Slicing strategy to adaptively divide the video into multiple groups of pictures(GOP) based on the dynamic level of the video in order to handle large camera motion and non-rigid movements. Finally, we propose quantization-aware fine-tuning to avoid performance reduction after quantization and utilize image codecs to compress Gaussians to achieve a compact representation.
Experiments on the Bunny and UVG datasets confirm that our method converges much faster than existing methods and also has 10x faster decoding speed compared to other methods. Our method has comparable performance in the video interpolation task to SOTA and attains better video compression performance than NeRV. 

\end{abstract}




\begin{teaserfigure}
  \includegraphics[width=\textwidth]{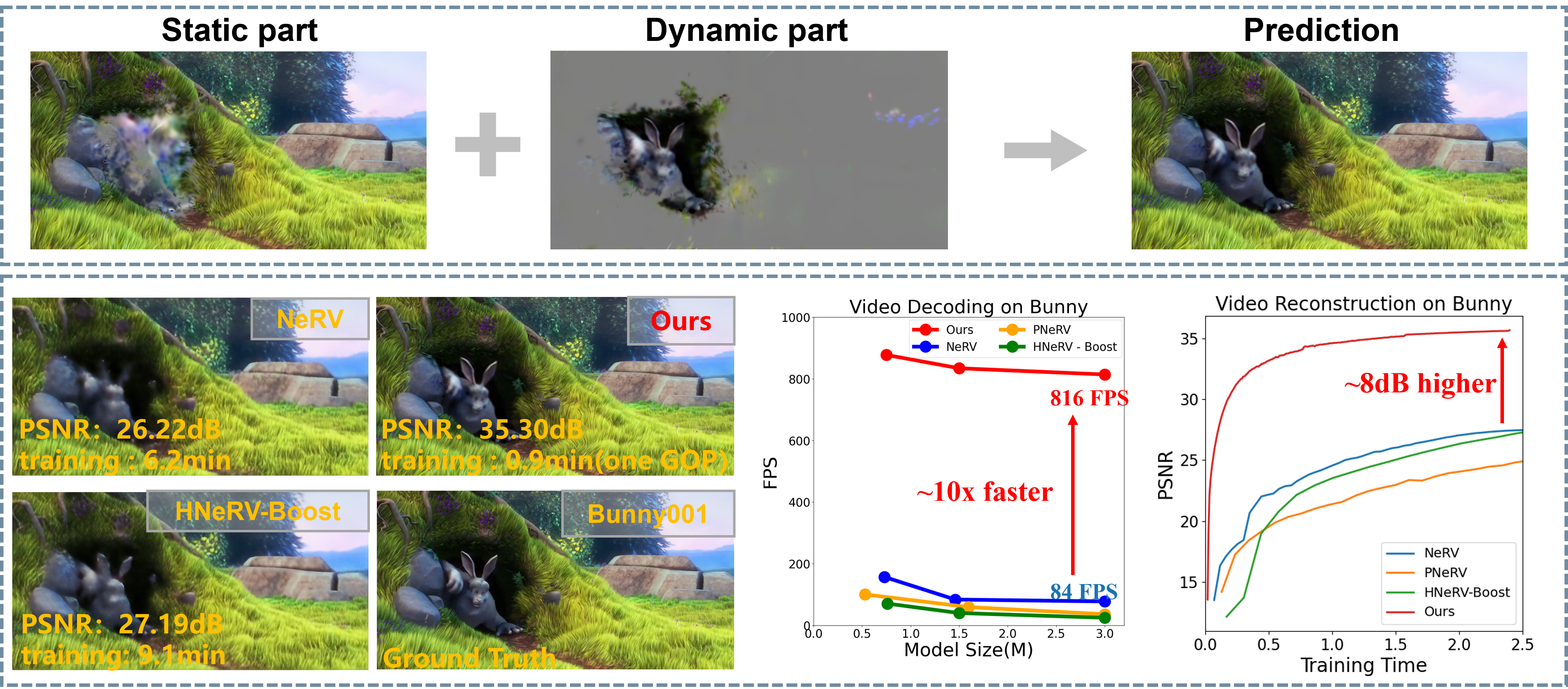}
  \caption{(Top) Our proposed method decomposes the video into dynamic and static part, depending on the dynamic indicator of 2D Gaussians. (Bottom) The visualization results of video reconstruction on Bunny and the video decoding and reconstuction result on Bunny.}
  \Description{teaser}
  \label{fig:teaser}
\end{teaserfigure}


\maketitle

\section{Introduction}
%背景
Recently, implicit neural representation (INR) has been proposed to represent discrete signals as continuous neural networks which uses a parameterized neural network to map coordinates to target outputs, such as rgb and density. INR has been used to represent a variety of signals, such as pictures~\cite{sitzmann2019siren}, videos~\cite{chen2021nerv}, static scenes ~\cite{mildenhall2020nerf}, dynamic scenes~\cite{pumarola2020dnerf}, etc.

Compared with traditional video codecs such as h.264 and HEVC, the implicit representation of video supports downstream tasks such as video interpolation and video inpainting. In addition, the neural implicit representation converts the video compression problem into a neural network model compression problem, greatly simplifying the encoding process compared to the complex pipeline of existing traditional video codecs. The implicit representation of video supports the individual decoding of any frame, while the traditional codecs must refer to the keyframe to decode following frames.

\begin{figure}[t]
\centering
\includegraphics[width=\linewidth]{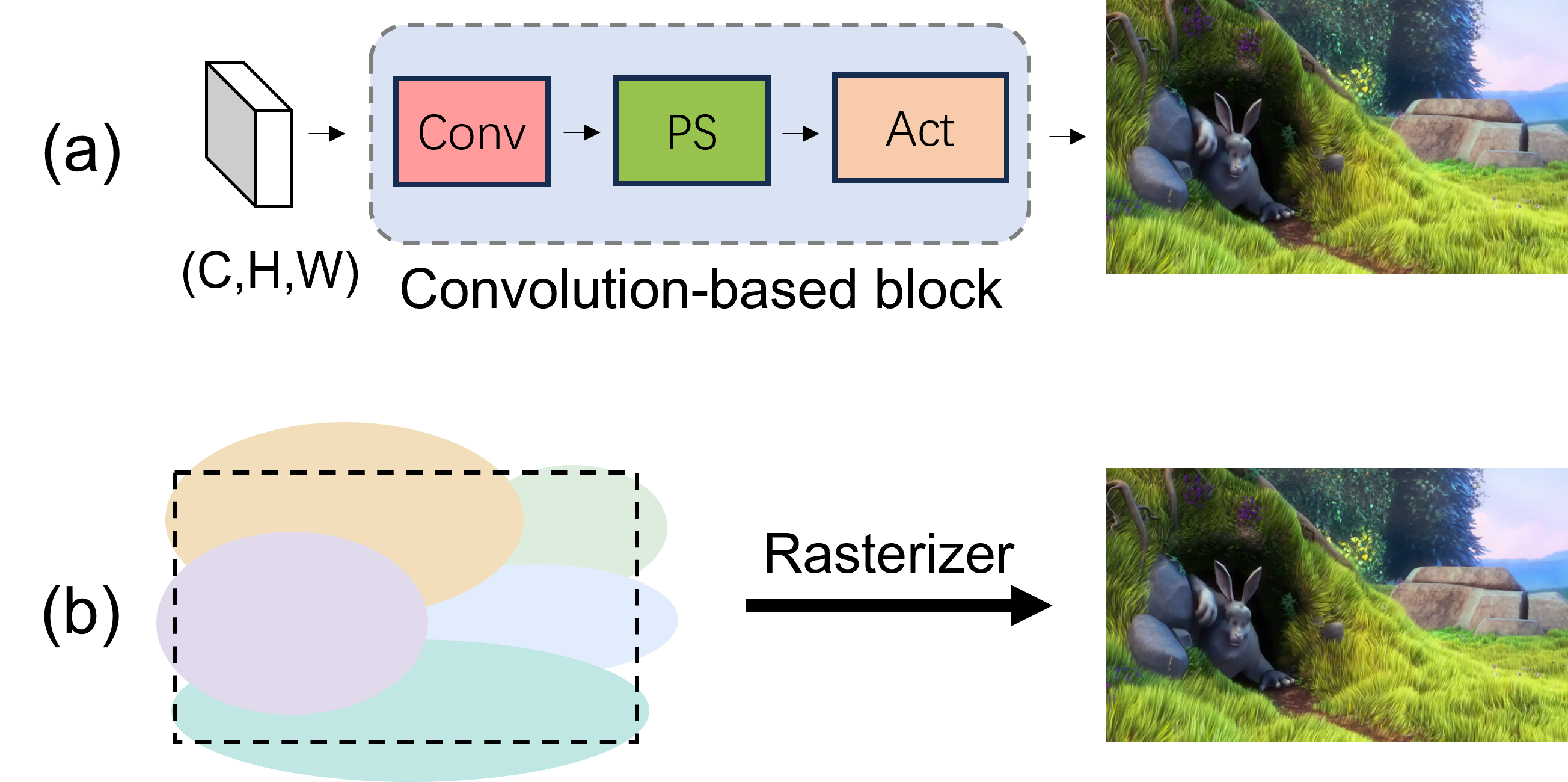}
\caption{The decoding process of (a) convolution-based video representation, (b) the proposed  representation.}
\label{fig:Two-method-compare.}
\Description{Two method compare.}
\end{figure}

% 引全
Due to these advantages, implicit representation of video~(NeRV) has become an emerging and promising representation for videos. There are two main types of existing convolution-based video representations, index-based~\cite{chen2021nerv,li2022enerv,lee2023ffnerv,yan2024ds,bai2023ps,zhang2024boosting,kwan2023hinerv} and frame-based~\cite{chen2023hnerv,zhao2023dnerv,zhao2024pnerv,wu2024qsnerv,he2023towards,saethre2024combining,xu2024vq,kim2024snerv}.
The index-based methods only utilize the index of each frame, while the frame-based methods combine with the frame embedding obtained from the encoder.
However, both methods share a similar decoder structure, known as NeRV block,as shown in Figure~\ref{fig:Two-method-compare.}. A typical NeRV block-based model is composed of above 5 convolutional layers and the computational complexity of convolution operations is high, which leads to slow training and inference speeds.  Experiments conducted on RTX 4090 demonstrate that existing methods only achieve about 30 fps~\cite{wu2024qsnerv} for 1920x960 videos, which is much lower than the frame rate of video~(i.e., 120 fps).

Inspired by the newly emerged image representation method GaussImage~\cite{zhang2024gaussianimage}, which represents image with 2D Gaussians and achieves superior decoding and compression performance, we propose GSVR, a 2D Gaussian-based video representation with hybrid deformation field.

However, in the video, the camera motion and object motion are tightly coupled, which poses significant challenges on the deformation field to model the dynamic of video. Thus, we propose a hybrid deformation field, which adopts the tri-plane grid to model camera motion based on the insight that adjacent Gaussians usually have similar spatio-temporal information. Then, a linear polynomial basis is introduced to model the motion of high dynamic objects. Moreover, we utilize a dynamic indicator to help each gaussian adaptively select its motion mode.
Note that our hybrid deformation field gets rid of the need for any neural networks, thereby achieving superior fast training and inference speed.

Besides, the deformation field also suffer from poor dynamic reconstruction in long video sequences due to the large camera motion and violent non-rigid motion of objects~\cite{shaw2025swings}.
Inspired from the group of pictures~(GOP) structure in traditional video codecs, we try to divide the video into multiple GOPs and reconstruct each GOP through our proposed GSVR.

Considering that the intensity of camera and object motion varies in different GOPs, we propose a dynamic-aware time slicing strategy to divide the video basing on the dynamic level of video. 

To reduce storage, we first quantize Gaussian attributes, then quantization-aware fine-tuning is applied to avoid performance decreasing. After that, disordered Gaussian attributes are mapped into 2D grids. These 2D grids and tri-plane grids are compressed by image codecs~(e.g., PNG, JPEG-XL).

In summary,our contributions are as follows:
\begin{itemize}
    \item We propose a novel 2D Gaussian-based video representation, GSVR,  which utilize a hybrid deformation field to model the complex motion of camera and objects in the video. By introducing a learnable dynamic indictor, the hybrid deformation field have the ability to separate the dynamic and static elements of the video.
    
    \item We propose a dynamic-aware time slicing strategy to segment a video into multiple GOPs basing on the dynamic level of the video, in order to handle camera motion and large non-rigid movements. After Gaussian attributes quantization, we apply quantization-aware fine-tuning to avoid performance decreasing and then map disordered Gaussian attributes as 2D grids to utilize image codecs to compress.
    
    \item We conduct experiments on two datasets, Bunny and UVG. Experiment results demonstrate that GSVR converge faster comparing to existing NeRV methods while with a significantly faster decoding speed, as shown in Figure~\ref{fig:teaser}. Our method achieves state of the art video interpolation performance and better compression performance than NeRV. 

\end{itemize}
\section{Related Work}

\subsection{Gaussian Image Representation}

3D Gaussian splatting is proposed as a novel 3D scene reconstruction method~\cite{kerbl20233dgs}, achieving real time rendering and state of the art scene quality. MiraGe~\cite{waczynska2024mirage} emphasizes image editing and directly adopts the 3DGS framework. Instead, some studies reduce the dimensionality of Gaussian attributes, transforming 3D ellipsoids into 2D ellipses and rendering images with a custom rasterizer to achieve significant fast decoding speed. GaussianImage~\cite{zhang2024gaussianimage} focuses on decoding speed by removing the depth-sorting process and opacity attribute required in 3DGS.  Image-GS~\cite{zhang2024image} proposes a method for adaptively generating Gaussians on the image plane and accelerates inference through hierarchical spatial partitioning. The application of 2D Gaussians in image representation demonstrates its advantages and potential in image compression, image decoding and image editing. These works prompt us to consider whether we can utilize 2D Gaussians to represent videos, enabling real-time decoding and other downstream tasks related to videos.

\subsection{Neural Representation for Videos}
Implicit Neural Representation can be used to represent various complex signals. SIREN~\cite{sitzmann2019siren} uses a Multi-Layer Perceptron~(MLP) to map the temporal information and pixel coordinates to the corresponding color values at the respective time instants and pixel positions. NeRV~\cite{chen2021nerv} replaces the MLPs with convolutional layers and upsampling layers. It only takes the temporal coordinates as input, avoiding repeated sampling at the pixel level and improving the decoding speed compared to previous methods.
Subsequent work of NeRV mainly focuses on improving video reconstruction quality, compression efficiency, and rarely pays attention to how to improve decoding speed. 
ENeRV~\cite{li2022enerv} introduces spatial information, reducing redundant model parameters while preserving the representational ability.
HNeRV~\cite{chen2023hnerv} reconstructs the target frames through the frame embeddings obtained by the encoder, which improves the regression capacity and internal generalization for video interpolation.
DNeRV~\cite{zhao2023dnerv} and PNeRV~\cite{zhao2024pnerv} introduce differential frame information and multi-scale information with specifically-designed fusion modules to improve the video quality. DS-NeRV~\cite{yan2024ds} compresses redundant static information by decomposing the video into static and dynamic latent codes and fuse these codes by an attention-based fusion decoder. BoostingNeRV~\cite{zhang2024boosting} utilizes a temporal-aware affine transform module to effectively align the intermediate features with the target frames.
% TODO
While these methods introduce complex network architectures to improve video quality, this improvement comes at the cost of compromised decoding speed due to the increased computational overhead.

With the successful application of 3D Gaussian in dynamic scenes, some methods have focused on representing videos through 3D Gaussians to achieve various downstream video tasks. Splatter a Video~\cite{sun2024splatter} enables video edition, dense tracking, and consistent depth generation by treating videos as the projection results of 3DGS in a fixed perspective within a 3D space. However, directly recovering 3DGS from monocular videos is a challenging problem and this approach requires accurate prior supervision. GaussianVideo~\cite{bond2025gaussianvideo} adapts the 3DGS formulation to model video data and model camera motion, enabling video stylization by propagate the change across the entire video. However, the video representation based on 3D Gaussians has a large number of parameters, which is not conducive to video compression.

Concurrent works D2GV~\cite{liu2025d2gv} and  GaussianVideo~\cite{lee2025gaussianvideo} utilize the GaussImage framework with deformation field to model videos. These works adopt pure MLP or multiplane-based encoder-decoder structure as the deformation field. However, due to the lack of explicit motion modeling, these methods are prone to failure when representing areas with rapid motion. Besides, due to high computational over-head, MLP will reduce the decoding speed significantly. Instead, we noticed the poor performance of a simple deformation field in representing regions with vigorous motion and propose the hybrid deformation field to solve this problem. Besides, our method completely get rid of MLP and achieve 800+ FPS decoding speed.

\subsection{Dynamic 3D Gaussian Splatting}
Some works extends 3DGS to represent dynamic scenes. There are two main methods, deformation field or time conditional probability. The deformation field methods~\cite{yang2024deformable, wu20244d, li2024spacetime,wan2024superpoint,huang2024sc,liang2023gaufre} employ various forms of deformation fields, such as MLPs, explicit grids, or polynomial motion. It takes time and the Gaussian position coordinates as input and outputs the Gaussian deformation at different timestamps. This method separates 3D geometry and appearance from motion but struggles to handle rapid movements. It also performs poorly when dealing with long-term sequences.

Other methods~\cite{yang2023gs4d,duan20244d} treat the spatiotemporal scene as a unified 4D Gaussian representation, where the 3D Gaussians at a given moment are considered as a conditional distribution of the 4D Gaussian representation under the condition of timestamp. This method couples space and time, lacks proper motion modeling, and may forcibly fit motion through opacity adjustments, potentially leading to poor generalization performance.

The success of 3D Gaussians combined with deformation fields in representing dynamic scenes inspires us to choose the deformation field scheme for effectively representing videos.

\begin{figure*}[ht] % 't' for top
    \centering
      \includegraphics[width=\textwidth]{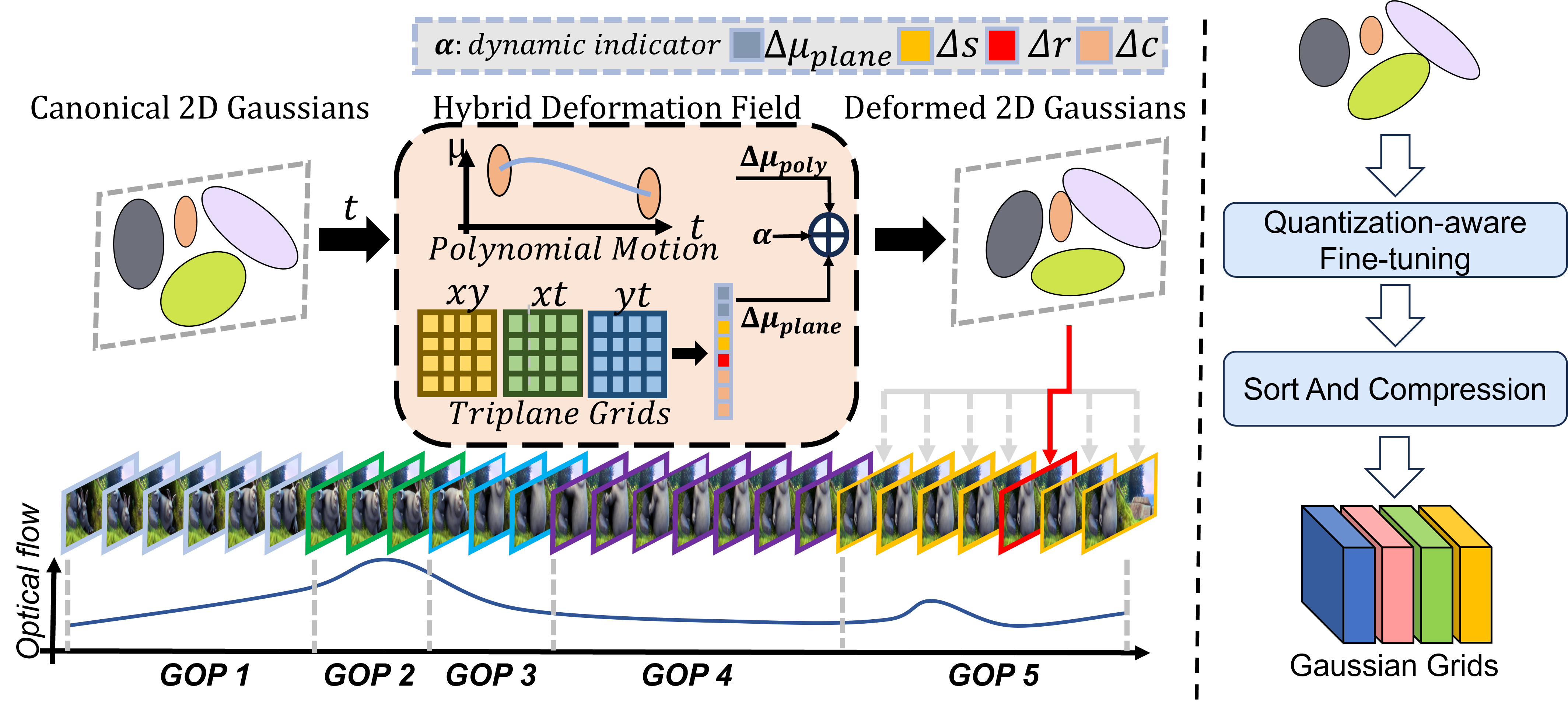} %
    \caption{GSVR framework overview. Our method uses tri-plane grids and polymonical motion to model the deformation of 2D Gaussians. The entire video is divided into multiple GOPs based on optical flow, and each GOP is represented by independent 2D Gaussians and deformation field. The canonial 2D Gaussians are compressed by quantization-aware fine-tuning and image encoder.}
    % A dynamic value \(\alpha\) is used for each gaussian to decide which motion pattern to choose.
    \label{fig:method}
    \Description{GS-NeRV framework overview.}
\end{figure*}

\section{Method}
% 介绍方法
Existing convolution-based video representation methods are inherently burdened by high computational complexity. To solve this issue, we introduce 2D Gaussian as the fundamental primitive for video representation. The deformation of 2D Gaussian is modeled using a hybrid deformation field that integrates tri-plane grids with polynomial motion functions to deal with the coupling of the camera motion and objects motion in the video. As shown in Figure~\ref{fig:method}, our framework includes Canonical 2D Gaussians $G$ and the deformation field network $F$.
Due to the difficulty for deformation filed to model long sequences, we divide the video into multiple GOPs. As the imbalance of motion level distribution cross the entire video sequences, a Dynamic-aware time-slicing strategy is proposed.
Finally, we employ quantization-aware fine-tuning to achieve compact representation.

\subsection{Canonical 2D Gaussian}

We adopt GaussImage~\cite{zhang2024gaussianimage} framework as the canonical 2D Gaussian representation. Each 2D Gaussian $\mathcal{G}$ has four attributes $G = \{\mu, s, \theta, c\}$, representing position $\mu$, scale $s$, rotation $\theta$, and color $c$.

For the case of 2D dimension, the covariance matrix can be formulated as a scale matrix $S$ and a rotation matrix $R$.
\begin{equation}
\Sigma = R S S^T R^T
\end{equation}%
where the rotation matrix R and the scaling matrix S are
\begin{equation}
R = \left[ 
\begin{array}{cc}
\cos(\theta) & -\sin(\theta) \\
\sin(\theta) & \cos(\theta)
\end{array} 
\right],\quad  S = \left[ 
\begin{array}{cc}
s_x & 0 \\
0 & s_y
\end{array} 
\right]
\end{equation}

Finally, the Gaussian distribution is represented by the follows:

\begin{equation}
f(\mathbf{x}) = \exp\left(-\frac{1}{2} (\mathbf{x} - \boldsymbol{\mu})^T \Sigma^{-1} (\mathbf{x} - \boldsymbol{\mu})\right)
\end{equation}where $\mu$ is the position of the Gaussian distribution.

\subsection{Hybrid Deformation Field}
The dynamic 3D Gaussian methods use deformation field to represent dynamic scenes, which can be MLP~\cite{yang2024deformable}, HexPlane~\cite{wu20244d} or hash grids~\cite{sun20243dgstream}. 
As a MLP-based deformation field has high computation overhead, it will effect inference speed significantly. We adopt explicit plane grids as the deformation field for lower computation demand.
Existing grid-based deformation field methods typically have a tiny MLP decoder to decode the grid feature into Gaussian deformations. However, we observed that MLP not only decrease the inference speed but also has no benefit to the video reconstruction quality. Thus, we remove the MLP decoder to achieve superior FPS.

We adopt tri-plane grids to learn the spatio-temporal relationship, due to adjacent Gaussians usually have similar spatial and temporal information. Specifically, our tri-plane grids contain 3 single-resolution plane modules. Each plane module is defined by $R(i, j) \in \mathbb{R}^{C\times N_i\times N_j}$, where $N$ indicates the resolution of the plane. Since we removed the tiny MLP decoder, $C$ is the same dim as the gaussian attribute offsets and we directly obtain the gaussian offset from the feature channel $C$. 
% $C$ is the same dim as the gaussian attribute offsets. 
The first two dimensions of the tri-plane grids feature represent position offset, the third and fourth dimensions represent scale offset, the fifth dimension represent rotation offset and the sixth to eighth dimensions represent color offset.

The position of 2D Gaussians $\mu=(x,y)$ and the timestamp $t$ are input into the plane module to obtain the offset of gaussian attribute.

\begin{equation}
% &表示对齐位置
\begin{aligned}
(\Delta\mu_{plane}^t,\Delta s^t,\Delta \theta^t,\Delta c^t) &= \prod \text{interp}(R(i,j)), \\
 (i,j) &\in \{(x,y),(x,t), (y,t)\},
\end{aligned}
\end{equation}
where '$interp$' denotes the bilinear interpolation for querying the voxel features located at 4 vertices of the grid.
Then, the deformed Gaussian attributes $\mathcal{G}_t=\{\mu^t, s^t, \theta^t, c^t\}$ at the current moment $t$ is obtained by the following formulas.

\begin{align}
s^t &= \exp(s) + \Delta s^t \\
\theta^t &= \pi \times \tanh(\theta) + \Delta \theta^t \\
c^t &= c + \Delta c^t \\
\mu^t &= \mu + \Delta\mu_{\text{plane}}^t
\end{align}where $exp$ represents exponential activation function to achieve fast convergence.

However, we observed that tri-plane grids exhibit poor performance when reconstructing high dynamic objects within video sequences, often leading to blurriness as shown in Figure~\ref{fig:ablation_on_deform}. The main reasons are as follows: (1) 3D dynamic scenes reconstruction task typically utilize camera parameters as input, while
for video reconstruction task, the camera parameters are unknown. Moreover,
the tightly coupling between camera motion and object moving creates complex motion patterns, which imposes greatly challenges on deformation field. (2) The deformation of canonical 2D Gaussians is interpolated from limited resolution of feature planes. However, the nearby Gaussians may have completely different motions, for example, the vicinity of the Gaussian of a dynamic object's edge is all static Gaussian. This phenomenon makes tri-plane grids fails to capture fine details of moving objects.

An intuitive approach to solve this problem is to increase grid resolution, however, this significantly increases model size. Instead, we introduced linear polynomial basis to model the motion of high dynamic objects, which can be represented by:
\begin{equation}
 \Delta\mu_{poly}^t = a_nt^n + a_{n - 1}t^{n - 1} + \cdots + a_1t + a_0,
\end{equation}
where $n$ is a non-negative integer, $a_i, i\in{\{1,2,\cdots,n\}}$ are the polynomial coefficients. In this paper, we set the $n=2$, and thus $a_0$, $a_1$, and $a_2$ are learnable parameter.

In our hybrid deformation field, tri-plane grids model the camera motion, static background, and slow moving objects while the linear polynomial basis models the high dynamic objects within a video. Furthermore, we also introduce a learnable dynamic indictor $\alpha$ for each Gaussian. Then, the position offset of Gaussian is obtained by fusing the prediction of tri-plane grids and the linear polynomial basis through $\alpha$ as follows:
\begin{equation}
\mu^t =  \mu + \alpha \cdot\Delta\mu_{poly}^t + (1-\alpha) \cdot \Delta\mu_{plane}^t 
\end{equation}%
where $\Delta\mu_{poly}$ is the predicted offset of the linear polynomial basis and $\Delta\mu_{plane}$ is the predicted offset of tri-plane grids. 
Note that our hybrid deformation field has a capability to separate background and dynamic elements during the learning process through dynamic indictor $\alpha$. Figure~\ref{fig:teaser} (Top) demonstrates that our method effectively distinguishes between dynamic and static elements. The rabbit part has a higher dynamic value, while the background part has a lower one.

Finally, we obtain the deformed 2D Gaussian $G^t = \{\mu^t, s^t, \theta^t, c^t\}$. 
These deformed 2D Gaussians in the screen space are used to rasterize the frame by weighted mixing as follows:
\begin{equation}
C_i  = \sum_{n \in \mathbb{N}} c_n^t  \cdot \exp(-\sigma_n), \quad
\sigma_n = \frac{1}{2} \delta_n^T (\Sigma^t)^{-1}\delta_n
\end{equation}%
where $\delta_n$ represents the distance with the pixel $i$ in the screen space, $N$ is the number of gaussians covering this pixel. As a result, our GSVR representation without any MLPs and can achieve more than 800+ FPS.

\subsection{Dynamic-aware Time Slicing}
% 改完trad，不是restore.
It is difficult to maintain a high reconstruction quality for long video sequences with a shared canonical space and hybrid deformation field.
Traditional video encoders utlize a GOP structure to encode video. Similar to that, we segment the video to handle camera movement and large non-rigid motions. We observe that the motion level in video varies significantly across different time segments. A fixed GOP length can lead to large disparities in PSNR between different GOPs.
Inspired by ~\cite{shaw2025swings}, we propose a Dynamic-aware Time Slicing strategy, Specifically, we use a pre-trained optical flow estimation model RAFT~\cite{2020RAFT} to evaluate the motion of each frame and adaptively segment videos into multiple GOPs according to the estimated motion level. Specifically, we use the average of the absolute values of the optical flow map of each frame as a measure of the degree of motion $D$.
\begin{align}
D = \frac{1}{HW} \sum_{i=1}^{H} \sum_{j=1}^{W}|flow\_image[i][j]|.
\end{align}%
Then, we accumulate the degree of motion $D$ of each frame from zero. After it exceeds a certain threshold, we compose one GOP with these frames and continue this process until the last frame of the video. Each GOP is represented by independent 2D Gaussians and deformation field.

\subsection{Gaussian Compression}

% \begin{figure}[t]
% \centering
% \includegraphics[width=\linewidth]{fig/2dgs-grid.png}
% \caption{Representing 2D Gaussian with 2D attribute Grids.}
% \label{fig:gs_grid}
% \Description{gs-grid.}
% \end{figure}

The trained Gaussian attributes occupy significant storage space, as each original attribute requires 32 bits for storage. To address this issue, we propose a quantization-aware fine-tuning strategy. 
For fine-tuning, we utilize quantization-aware training with Min-Max quantization. In the forward pass, the quantization of gaussian attributes is simulated using a rounding operation considering the number of bits.

On the other side, the unstructured list of Gaussian attributes contains substantial redundancy, as spatially adjacent positions often exhibit high attribute similarity, while the random-order list stores each gaussian independently without exploiting neighboring relationships. To resolve this problem, we propose mapping the Gaussian list into a 2D grid and compressing redundancies between adjacent attributes using an image encoder, similar to ~\cite{morgenstern2024compact}.
 
Specifically, we firstly use 16 bits to quantize the position and color, 8 bits to quantize other attributes. Then, the quantized GSVR is finetuned to improve the reconstruction performance while maintain a compact representation. After finetuning, we map all Gaussian attributes to 2D grid. Then, image encoder (e.g., JPEG-XL, PNG) is applied to further compress gaussian attributes. As for tri-plane grids, we use the 16 bits PNG image format for compression. 
As a result, we compressed Bunny's 3M parameters to 3.1 MB with a compression ratio of 3.68.

\begin{figure*}[t]
\centering
\includegraphics[width=\linewidth]{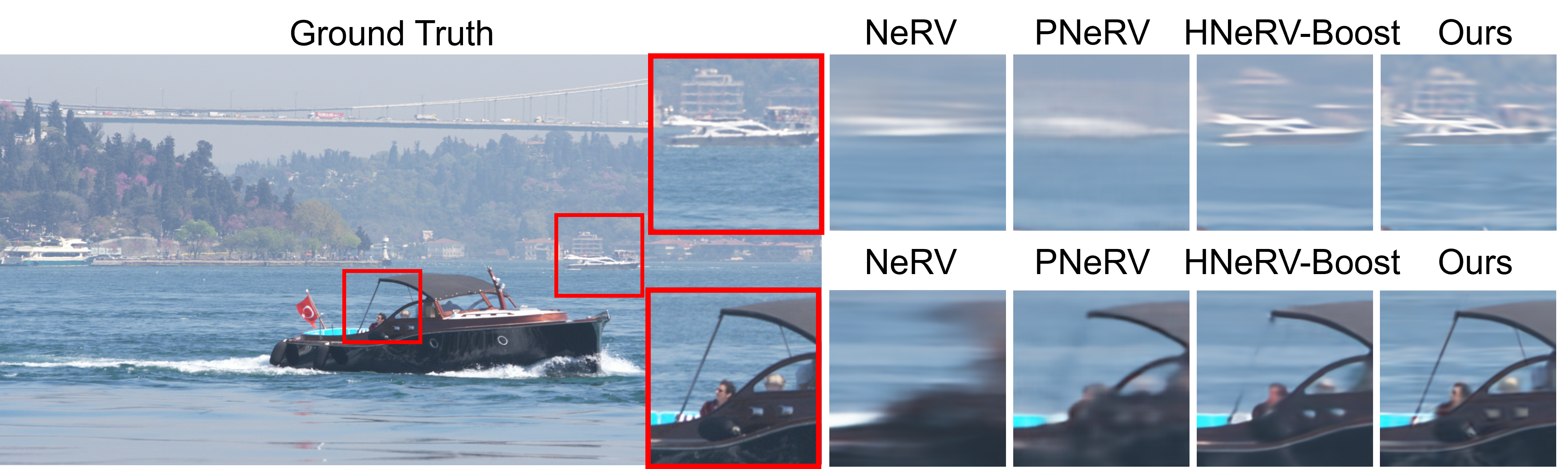}
\caption{Video reconstruction result on UVG. Training time is 1.}
\label{fig:Video_reconstruction}
\Description{Video_construction.}
\end{figure*}
\begin{table*}[!t]
  \caption{Video reconstruction results on UVG, PSNR reported. Training time is 1.}
\label{tab:video_reconstruction_uvg}
  \begin{tabular}{c|c c c c c c c c}
    \toprule
    960x1920 & beauty & bosph & honey & jockey & ready & shake & yacht & avg. \\ 
    \midrule
    NeRV & 27.71 &27.26 & 32.68 & 21.48& 17.59 & 27.04 & 22.89 &25.23 \\ 
    PNeRV & 25.99 & 26.07 & 29.92 & 20.35 & 16.76 & 26.81 &21.60 &23.93 \\ 
    HNeRV-Boost & \textbf{31.53} &30.41 &37.27 &\textbf{26.54}& 22.16& 30.92 & 25.30 &29.16\\ \hline
    Ours &31.39&\textbf{33.83}   & \textbf{38.10} & 24.30& \textbf{23.99} & \textbf{32.34} &\textbf{27.59} & \textbf{30.22} \\ 
    \bottomrule
  \end{tabular}
\end{table*}
\begin{table}[ht]
  \caption{Video reconstruction on Bunny.}
  \label{tab:reconstruction_on_bunny}
  \begin{tabular}{c|c c c c c}
    \toprule
    Training Time&0.5 & 1.0 & 1.5 & 2.0 \\ 
    \midrule
     NeRV &22.24&24.75&26.17&27.14\\
     PNeRV &20.34&22.07&23.36&24.25\\
    HNeRV-Boost&20.77&23.53&25.25&26.62\\ \hline
     Ours &\textbf{33.17}& \textbf{34.60}& \textbf{35.14} & \textbf{35.48}\\ 
    \bottomrule
    \end{tabular}
\end{table}

\begin{table}[ht]
  \caption{Video reconstruction on UVG.}
  \label{tab:reconstruction_on_bunny}
  \begin{tabular}{c|c c c c c c}
    \toprule
    Training Time&0.2 & 0.4 & 0.6 & 0.8 & 1.0 \\ 
    \midrule
     NeRV&21.18&23.24&24.39&25.09&25.23\\
     PNeRV&18.09&20.67&22.01&22.96&23.93\\
    HNeRV-Boost&20.19&25.50&27.50&28.60&29.16\\ \hline
     Ours &\textbf{28.63}& \textbf{29.39}& \textbf{29.79} &\textbf{30.06}& \textbf{30.22}\\ 
    \bottomrule
    \end{tabular}
\end{table}
\section{Experiments}
\subsection{Setup}
\subsubsection{Datasets}
We choose two datasets, the Big Buck Bunny~\cite{bunny} and UVG~\cite{uvg}. The Big Buck Bunny has 132 frames with a size of $720\times1280$. UVG has 7 videos with a size of $1080\times1920$ in 300 or 600 frames. Following regular setting, we crop Bunny to $640\times1280$ and UVG to $960\times1920$.
\subsubsection{Metrics}
We employ peak signal-to-noise ratio (\textbf{PSNR}) as a metric to evaluate video reconstruction quality, and bits per pixel (\textbf{bpp}) to evaluate video compression performance. We define \textbf{training time} as 
total training seconds divided by the total frame numbers~(i.e., time required for training each frame).
Since DS-NeRV~\cite{yan2024ds} does not release the code, we choose PNeRV~\cite{zhao2024pnerv}, HNeRV-Boost~\cite{zhang2024boosting} as the baseline models. We also choose NeRV~\cite{chen2021nerv} since this method has relatively fast decoding speed. Unless stated otherwise, all models are 3M parameters and all experiments are conducted on the RTX A6000. For other methods, we adjust the parameters to about 3M.

\begin{figure}
% \vspace{-0.6cm}
\centering
\includegraphics[width=\linewidth]{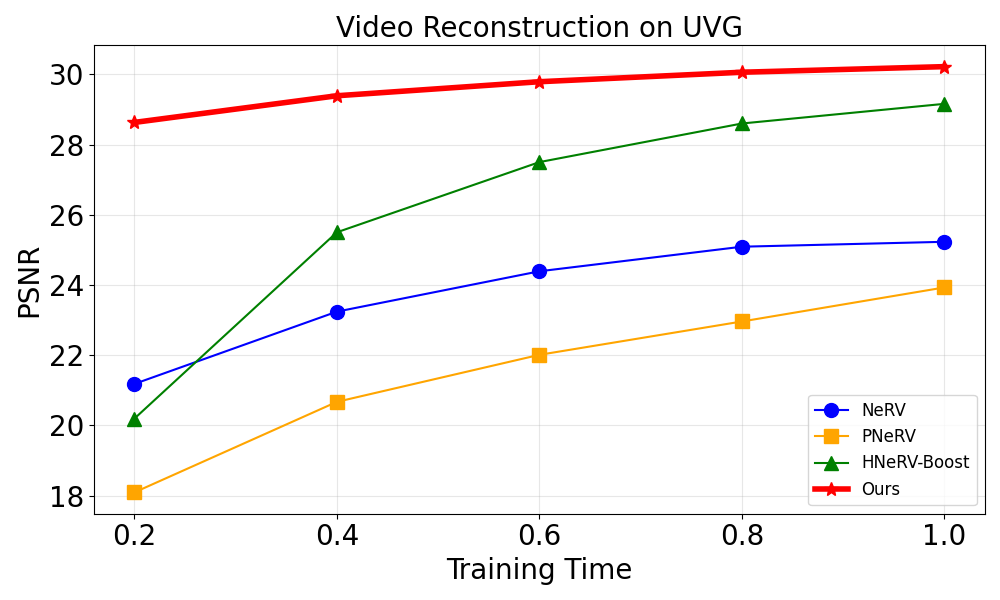}
\caption{Video reconstruction on UVG}
\label{fig:Video reconstruction on UVG.}
\Description{Video reconstruction on UVG}
\end{figure}

\subsubsection{Implementation Details}
The rasterizering process of 2D Gaussian is based on gsplat~\cite{ye2024gsplatopensourcelibrarygaussian}.
The position of Gaussians are randomly initialized between -1 and 1. The rotation is initialized as zero and scale is one. The color of Gaussians are randomly initialized between -0.5 and 0.5. 
The time dimension of the tri-plane is the number of the frames divide by 2. The resolution of xy plane is $32\times16$ and the feature channel is 8.
The learning rate of position is 0.0025. The learning rate for other Gaussian attributes and tri-planes is 0.01.
We only use L2 loss to supervise our method.
\begin{equation}
    L_2(y_i, \hat{y}_i) = \frac{1}{n} \sum_{i=1}^n \left( y_i - \hat{y}_i \right)^2
\end{equation}where $y_i$ represents the ground truth and $\hat{y}_i$ is the reconstructed frame.

\subsection{Video Reconstruction}
% 不同编码时间的对比 Bunny

Since the forward propagation speed varies across different models, it is more fair to compare video reconstruction performance in the same training time rather than epochs.
As shown in Figure~\ref{fig:teaser}, for the Bunny dataset, with the same training time, our method converges more quickly and achieves higher video quality. Specifically, under the same training time, the PSNR of our method is 8dB higher than other methods, as shown in Table~\ref{tab:reconstruction_on_bunny}. 
Our method still exhibits faster convergence speed for the UVG dataset, as shown in Figure~\ref{fig:Video reconstruction on UVG.} and Figure~\ref{fig:Video_reconstruction}.  

\subsection{Video Decoding}
We evaluate the decoding speed by calculating the average inference speed over 100 forward passes of the model. 
Our method achieves 10x higher FPS comparing to NeRV due to our efficient architecture, as demonstrated in Table~\ref{tab:decoding_comparison}. Note that the vanilla NeRV still cannot achieve real-time (>60 fps) forward inference for videos at a resolution of 1920x960 on A6000. However, our method far exceeds this requirements, and has the potential for real-time decoding on low-performance GPUs.
\begin{table}[t]
  \caption{FPS results on Bunny and UVG at 3M model size.}
  \label{tab:decoding_comparison}
  \begin{tabular}{c|c|c}
    \toprule
    Method & Bunny&UVG(Mean)\\ 
    \midrule
    NeRV& 84.55&48.26\\
    PNeRV&36.17&19.7\\
    HNeRV-Boost&24.80&16.44\\\hline
    Ours &\textbf{816.56}&\textbf{538.49}\\
    \bottomrule
  \end{tabular}
\end{table}

\subsection{Ablation Studies}
\subsubsection{Deformation field}
\begin{figure}[!t]
\vspace{-0.4cm}
\centering
\includegraphics[width=\linewidth]{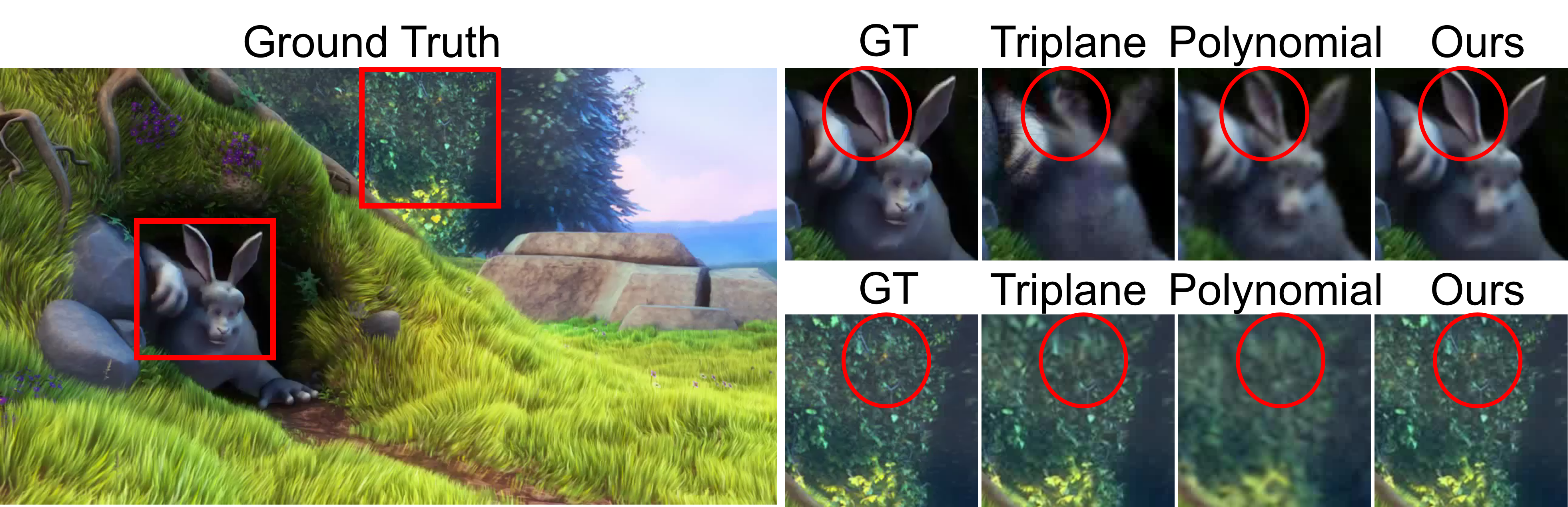}
\caption{Ablation experiment about motion pattern.}
\label{fig:ablation_on_deform}
\Description{Ablation experiment on Bunny about motion pattern.}
\end{figure}
Qualitative experiments showed that the tri-plane method struggled to learn object motion while easy to fit camera motion.
As shown in Figure~\ref{fig:ablation_on_deform}, using tri-plane only leads to insufficient fitting of moving objects, while the use of polynomial motion alone leads to insufficient fitting of static backgrounds.

While our proposed hybrid deformation field combine the advantages of tri-plane and polynomial basis to solve the camera motion and object motion in the video.
We select a threshold and use the Gaussians with dynamic indicator higher and lower than this threshold to render images separately. As shown in Figure~\ref{fig:teaser}, the background is modeled mainly by tri-plane motion, while the moving part of the rabbit is modeled by polynomial basis.

\subsubsection{Dynamic-aware Time Slicing}
Table~\ref{tab:psnr_gop_settings} demonstrates that dynamically adjusting the GOP length can better accommodate variations in video content, leading to superior reconstruction quality. All models are trained for 300 epochs, ensuring a fair comparison of PSNR across different GOP length settings. It can be seen that a GOP length that is too large or too small will lead to poor fitting in some videos such as honey and ready, while the adaptive GOP length achieves an acceptable trade-off. 

\begin{figure}[t]
\centering
\includegraphics[width=\linewidth]{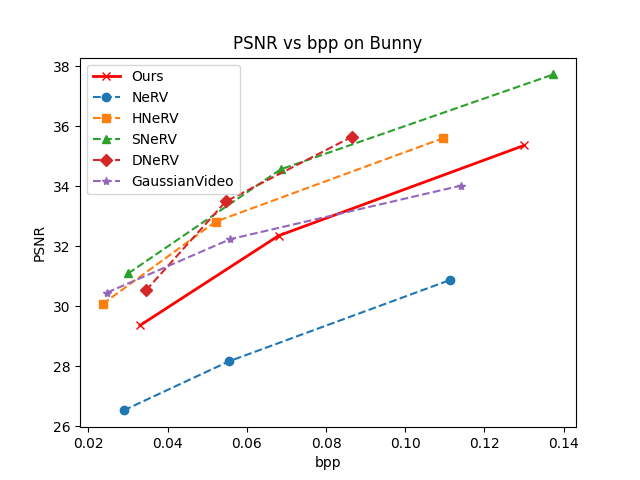}
\caption{Compression on Bunny, results from ~\cite{lee2025gaussianvideo}.}
\label{fig:bpp_psnr_bunny_gaussianvideo}
\Description{bpp_psnr_bunny_gaussianvideo}
\end{figure}

\subsection{Downstream Task}

\begin{figure*}[t]
\centering
\includegraphics[width=\linewidth]{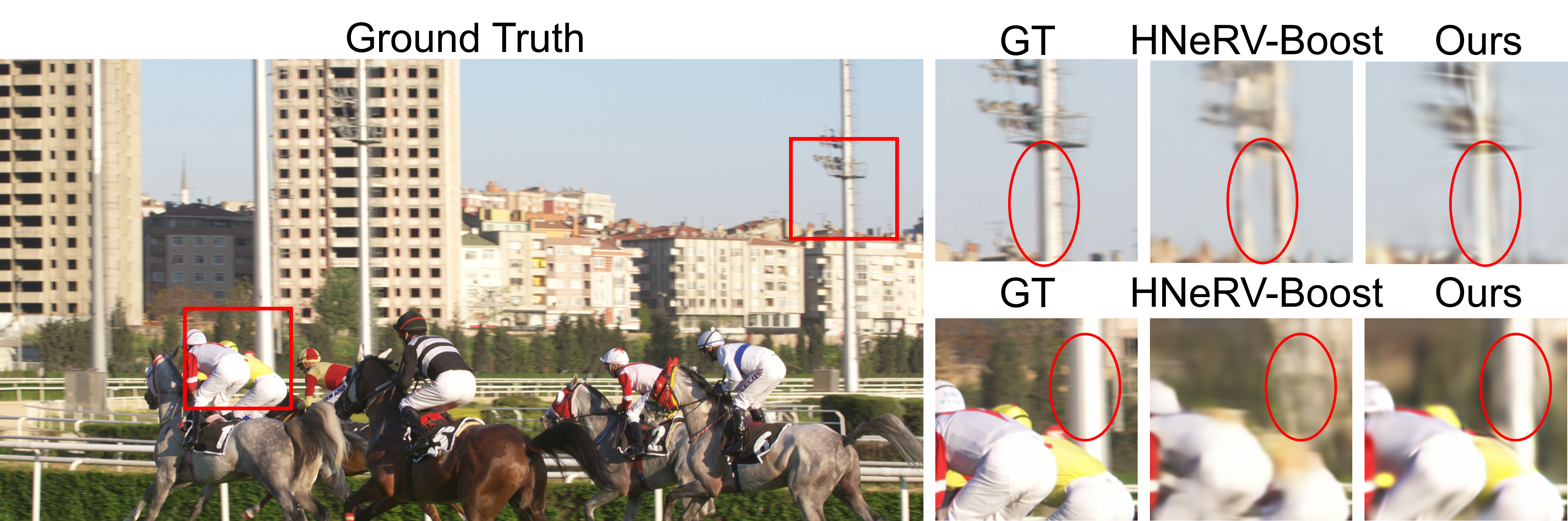}
\caption{Video interpolation result on UVG. Note that HNeRV-Boost experiences ghosting or disappearance of objects.}
\label{fig:video_interpolation}
\Description{video_interpolation.}
\end{figure*}
\begin{table*}[h]
    \caption{Video interpolation results on UVG with train/test split, PSNR($\uparrow$) reported.}
    \label{tab:interpolation}
        \begin{tabular}{c||ccccccc||c}
        \toprule
            video & Beauty & Bosph & Honey & Jockey & Ready & Shake & Yacht & avg. \\ 
            \midrule
            HNeRV-Boost~\cite{zhang2024boosting} & 32.53/31.36 & 32.51/32.51 & 38.25/\textbf{38.24} & 30.28/23.04 & 24.80/20.96 & 33.35/32.74 & 26.67/26.53 & 31.20/29.34\\ 
            Ours & 32.54/\textbf{31.57} & 32.51/\textbf{32.56} & 38.25/38.16 & 30.28/\textbf{23.63} & 24.49/\textbf{21.00} & 33.35/\textbf{32.81} & 26.68/\textbf{26.67} & 31.16/\textbf{29.49}\\ 
        \bottomrule
        \end{tabular}
\end{table*}
\begin{table*}[t]
  \caption{PSNR($\uparrow$) on UVG dataset with different GOP lengths for 300 training epochs.}
  \label{tab:psnr_gop_settings}
  \begin{tabular}{c|c c c c c c c c}
    \toprule
    GOP Length & beauty & bosph & honey & jockey & ready & shake & yacht & psnr\_mean \\ \hline
    \midrule
    15 & 32.45 & 32.79 & 31.50 & \textbf{27.59} & 23.85 & 30.91 & 27.58 & 29.52 \\ \hline
    30 & \textbf{32.61} & 34.47 & 34.86 & 27.12 & 23.52 & 32.44 & \textbf{27.90} & 30.42 \\ \hline
    60 & 32.47 & 34.30 & 37.25 & 26.64 & 23.03 & \textbf{33.07} & 27.52 & 30.61 \\ \hline
    100 & 32.33 & 33.76 & 38.13 & 26.44 & 22.59 & 33.06 & 27.16 & 30.50 \\ \hline
    Adaptive & 32.46 & \textbf{34.56} & \textbf{38.34} & 27.02 & \textbf{24.09} & 32.98 & 27.65 & \textbf{31.01} \\ 
    \bottomrule
  \end{tabular}
\end{table*}
\subsubsection{Video Interpolation} 

Due to the poor video interpolation performance of NeRV and PNeRV does not provide an implementation of video interpolation. We only compare the video interpolation performance with HNeRV-Boost. As shown in Table~\ref{tab:interpolation}, GSVR achieves the best video interpolation performance. Note that HNeRV-Boost requires test frames to form frame embeddings, but in practice, test frames can not be obtained.

Considering the faster convergence speed of our method, to ensure fairness, we maintain a consistent PSNR and model size in the training set and evaluate the PSNR in the test set. We use odd-numbered frames as the training set and even-numbered frames as the test set.

As shown in Figure~\ref{fig:video_interpolation}, as HNeRV-Boost decodes each frame independently basing on frame embedding, lacking temporal correlation between frames, the interpolation results show issues such as ghosting or disappearance of utility poles. In contrast, our method faithfully reconstructs a single utility pole.
Our method explicitly represents the movement of the scene and objects, and continuity with the preceding and subsequent frames is guaranteed during interpolation. 

\begin{table*}[t]
\centering
\caption{Compression results on Bunny after quantization and encoding.}
\label{tab:Compression_on_Bunny}
\begin{tabular}{l|l|l|l|l|l}
\hline
Methods & Param & PSNR & training time  &bits per param & bits per pixel \\
\hline
HNeRV-Boost & 3M & 29.11 & 9.0 minutes & 8.26 &  0.23 \\
Ours & 3M  &35.65  & 5.3 minutes & 9.01& 0.25 \\
\hline
HNeRV-Boost  & 4.5M & 29.49
 & 12.5 minutes &  8.23 &   0.34
 \\
Ours & 4.5M  & 38.29 &  12.6 minutes & 8.89& 0.37 \\
\bottomrule
\end{tabular}
\end{table*}

\subsubsection{Video Compression}
% As shown in Table~\ref{tab:Compression_on_Bunny}, our method achieve comparable compression of parameters with Boost-HNeRV after quantization and encoding. 
With short training time, our method perform superior rate distortion performance than BoostHNeRV. 

We also compare the compression results on Bunny with other NeRV variants methods and concurrent work GaussianVideo~\cite{lee2025gaussianvideo}, as shown in Figure~\ref{fig:bpp_psnr_bunny_gaussianvideo}. The experimental results establish that our method maintains competitive compression performance with contemporary NeRV variants, while achieving significantly better rate distortion performance compared to vanilla NeRV.

\section{Conclusion}
In this paper, we propose GSVR, a 2D Gaussian-based video representation
with hybrid deformation field. We utilize a dynamic-aware time slicing strategy to divide the video into different GOPs. We compress 2D Gaussians by quantization-aware fine-tuning and image encoder to achieve a compact representation. Due to the computational efficiency of 2D Gaussian blending, we achieve significantly faster decoding speeds than other NeRV-based methods and our method can also converge more quickly. 

Due to our GOP structure and low computational overhead, our method is more appropriate for video streaming and can be deployed on mobile devices such as smartphones, which will broaden the range of applications for existing neural implicit video representations.  Our work can inspire the use of neural implicit representations on mobile devices, in order to help low-end devices to decode high-resolution videos.

\bibliographystyle{ACM-Reference-Format}
\bibliography{sample-base}


\end{document}